\documentclass[conference]{IEEEtran}
\IEEEoverridecommandlockouts

\usepackage{cite}
\usepackage{amsmath,amssymb,amsfonts}
\usepackage{algorithmic}
\usepackage{graphicx}
\usepackage{textcomp}
\usepackage{xcolor}

\usepackage{multirow}
\usepackage{hyperref}
\usepackage{booktabs}
\usepackage{utfsym}
\usepackage{colortbl}
\usepackage{color}
\usepackage{float} 
\usepackage{bm}
\usepackage{subfigure}

\def\BibTeX{{\rm B\kern-.05em{\sc i\kern-.025em b}\kern-.08em
    T\kern-.1667em\lower.7ex\hbox{E}\kern-.125emX}}

\makeatletter
\newcommand{\linebreakand}{%
  \end{@IEEEauthorhalign}
  \hfill\mbox{}\par
  \mbox{}\hfill\begin{@IEEEauthorhalign}
}

\newcommand{\Rmnum}[1]{\expandafter\@slowromancap\romannumeral #1@}
\makeatother
    
\begin{document}

\title{
Enhancing Multimodal Sentiment Analysis for Missing Modality through Self-Distillation and Unified Modality Cross-Attention
}


\author{Yuzhe Weng\IEEEauthorrefmark{1}\IEEEauthorrefmark{2}, Haotian Wang\IEEEauthorrefmark{1}, Tian Gao\IEEEauthorrefmark{2}, Kewei Li\IEEEauthorrefmark{1}, Shutong Niu\IEEEauthorrefmark{1}, Jun Du\IEEEauthorrefmark{1}\IEEEauthorrefmark{4}\IEEEauthorrefmark{3}\thanks{\IEEEauthorrefmark{3} Corresponding author} \\
\IEEEauthorblockA{\IEEEauthorrefmark{1}NERC-SLIP, University of Science and Technology of China (USTC), Hefei, China}
\IEEEauthorblockA{\IEEEauthorrefmark{4}MoE Key Laboratory of Brain-inspired Intelligent Perception and Cognition, USTC, Hefei, China}
\IEEEauthorblockA{\IEEEauthorrefmark{2}iFlytek Research, Hefei, China}

\IEEEauthorblockA{
\{yzweng, az1522702192, keweili12, niust\}@mail.ustc.edu.cn, tiangao5@iflytek.com, jundu@ustc.edu.cn
}
}

\maketitle


\begin{abstract}
In multimodal sentiment analysis, collecting text data is often more challenging than video or audio due to higher annotation costs and inconsistent automatic speech recognition (ASR) quality. To address this challenge, our study has developed a robust model that effectively integrates multimodal sentiment information, even in the absence of text modality. Specifically, we have developed a Double-Flow Self-Distillation Framework, including Unified Modality Cross-Attention (UMCA) and Modality Imagination Autoencoder (MIA), which excels at processing both scenarios with complete modalities and those with missing text modality. In detail, when the text modality is missing, our framework uses the LLM-based model to simulate the text representation from the audio modality, while the MIA module supplements information from the other two modalities to make the simulated text representation similar to the real text representation. To further align the simulated and real representations, and to enable the model to capture the continuous nature of sample orders in sentiment valence regression tasks, we have also introduced the Rank-N Contrast (RNC) loss function. When testing on the CMU-MOSEI, our model achieved outstanding performance on MAE and significantly outperformed other models when text modality is missing. The code is available at: \href{https://github.com/WarmCongee/SDUMC}{https://github.com/WarmCongee/SDUMC}.
\end{abstract}
\section{Introduction}
\label{sec:intro}

Multimodal sentiment analysis has garnered significant attention \cite{lian2023mer, lian2024merbench, zhao2021emotion, asiya2022novel}, with numerous studies focusing on the representation \cite{yu2021learning, ma2023emotion2vec, chen2024vesper, wang2021distributed}, alignment \cite{fan2024atta, huang2024clip}, and fusion \cite{MULT, MFM, ALMT, AcFormer} of sentiment information from various modalities. The absence of certain modalities during training and inference significantly influences the alignment and fusion of multimodal information, prompting increased research efforts \cite{li2024correlation, zuo2023exploiting, zhao2024toward, dai2024study, 10447257} aimed at enhancing robustness under these conditions. Traditionally, research \cite{li2024correlation} has classified the absence of text, audio, and visual modalities as either complete or frame-level missing, often employing partial word masks to mimic text frame-level absence. However, this simulated absence differs markedly from typical real-world scenarios where text is either completely missing or substituted with low-quality ASR transcriptions. Given the text modality's superior performance \cite{ALMT} in certain tasks, previous studies \cite{li2024speech, shon21_interspeech} have relied on manually annotated transcriptions or the development of ASR/AVSR systems. We contend that it is unnecessary to reconstruct the final text when the text modality is missing. Instead, the model can simply use the hidden states derived from the audio or visual modalities to approximate the semantic space of the missing text modality.


Theoretically, textual information can be highly abstracted from audio and visual modalities. With advancements in Large Language Models (LLM), some studies \cite{ma2024SLAM-ASR, geng2024Chinese-LLM-ASR} have employed LLMs as decoders to address ASR tasks. For instance, SLAM-ASR \cite{ma2024SLAM-ASR} projects audio representations and inputs them with prompts into an LLM to generate ASR results. Consequently, it is natural to infer that simulated text representations can be derived from the projected audio representations via LLMs, facilitating unified multimodal sentiment analysis.

\begin{figure*}[ht]
\centering
\includegraphics[width=1.0\textwidth]{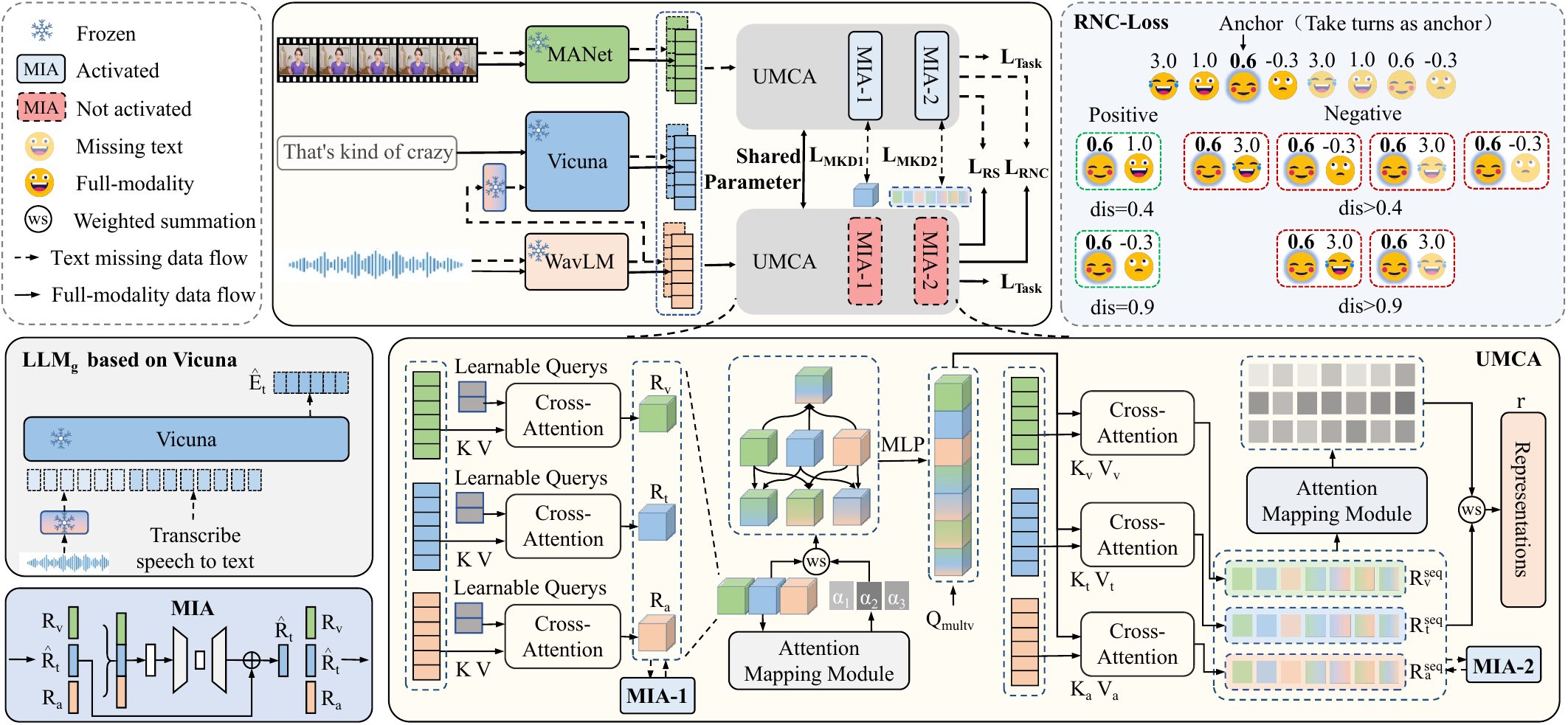}
\caption{Overall architecture of Double-Flow Self-Distillation Framework for modality missing. In the figure, the middle part of the upper row shows the training process of the framework. The lower row shows the specific network structure of each module.} 
\label{Figure 1.}
\end{figure*}

Our research proposes a Double-Flow Self-Distillation Framework, which includes the Modality Imagination Autoencoder (MIA) and Unified Modality Cross-Attention (UMCA). This unified model can address both complete and text-missing multimodal sentiment analysis scenarios by simply activating or deactivating the MIA during inference. The main contributions of our work are:
(1) We design UMCA for comprehensive multimodal fusion to extract sentiment-related information.
(2) Our method uses the LLM-based model Vicuna \cite{vicuna2023} to generate simulated text representations from audio representations' projection when text modality is missing, with MIA complementing the missing information.
(3) We train a unified multimodal sentiment analysis network using self-distillation combined with MIA activation and deactivation while achieving distance-space alignment of complete and missing modality representations through MKD Loss and RNC Loss\cite{zha2024RNC}. 
(4) Our method achieves optimal performance on the CMU-MOSEI \cite{CMU-MOSEI} across multiple metrics for complete modality inference, and significantly improving performance in text-missing scenarios compared to other methods.

\section{Method}
\label{sec:method}

In this section, we discuss our proposed Double-Flow Self-Distillation Framework for the multimodal sentiment analysis. The overall architecture of our system is depicted in Fig. \ref{Figure 1.}.

\subsection{Unified Modality Cross-Attention}
\label{ssec:Unified Modality Cross-Attention}

As shown in the bottom right corner of the Fig. \ref{Figure 1.}, the UMCA model is divided into two main stages. And we set up the cross-attention module, which is applied in both stages. The formulas are as follows: 
\begin{gather}
K_{m} = \mathrm{Tanh}(\bm{W}V_{m}+\bm{b})\\
R_{m} =\mathrm{Softmax}(\frac{QK_{m}^{\mathrm T}}{\tau})V_{m}
\end{gather}
$m \in \{\mathrm{a}, \mathrm{v}, \mathrm{t}\}$, where $\mathrm{a}$, $\mathrm{v}$, and $\mathrm{t}$ refer to the three modalities of audio, vision, and text. $\tau$ denotes the scaling parameter that ensures the stability of the gradient of softmax. $K_{m}$ and $V_{m}$ are the key and value sequences from the $m$ modality, $Q$ is the query involved in the cross-attention, and $R_{m}$ is the representation of the $m$ modality in the same dimension as $Q$.

The first stage is used to obtain the query embedding of the multi-view modalities combinations. First, each modality representation is mapped to the same dimension $E_{m} \in \mathbb{R}^{S\times D}$ by MLP, where $S$ represents the sequence length and $D$ denotes the dimension determined by a hyperparameter. We initialize the Gaussian-distributed learnable query ($Q_m$) for each modality during model initialization, where $Q_m \in \mathbb{R}^{\mathrm{S_1} \times D}$, with $S_1$ set to 1. For each modality, the original modal representations are mapped to the key ($K_m$) and value ($V_m$). According to the equation, we use the cross-attention module to obtain the representations $R_{m}$ for each modality. The Attention-guided Feature Gathering (AFG) module \cite{wang2023hierarchical} concatenates the three modal representations and feeds them into an MLP to obtain attention weights. These weights are used to compute the weighted sums of unimodal, bimodal, and trimodal representations, producing a multi-view query ($Q_{\mathrm{multv}}$) with seven combinations. For instance, in a bimodal scenario, we set the unused modality's weight to zero while maintaining the weights of the other two modalities. This results in a weighted sum that forms the bimodal query.

In the second stage, the multi-view query obtained in the previous stage is used as the query ($Q_{\mathrm{multv}}$). The original modal representations are mapped as key ($K_m$) and value ($V_m$). The modal representations $R_{m}^{\mathrm{seq}}$ are obtained by cross-attention, and the final representation $r$ is obtained by attention weighted summing $R_{m}^{\mathrm{seq}}$ by AFG module, and regression is performed to obtain the sentiment valence.
\subsection{Missing Modality Imagination}
\label{ssec:Modal Imagination}
\subsubsection{LLM-based Text Representation Simulation}
\label{sssec:LLM Decoding Audio Representation}

SLAM-ASR \cite{ma2024SLAM-ASR} uses WavLM as an audio encoder and LLM-based model Vicuna \cite{vicuna2023} as a decoder. Then the method trained the projector between them and generated ASR results from the audio encoder representations directly using the generative power of the LLM. Based on this, we work on unifying the representation of text modality presence and missing. Definitions are as follows: 
\begin{gather}
E_{\mathrm{t}} = {\rm LLM_f}({\rm tokenizer}(T)) \\
\hat{E_{\mathrm{t}}} = {\rm LLM_g}({\rm cat}({\rm proj}(E_{\mathrm{a}}), {\rm tokenizer}(P)))
\end{gather}
In the definitions, $T$ denotes the text transcribed from video content. The symbol $E_{\mathrm{t}}$ refers to the real text representation obtained from the text modality. The function ${\rm LLM_f(\cdot)}$ is employed for the direct forward pass to obtain the hidden state. Additionally, the symbol $\hat{E_{\mathrm{t}}}$ denotes the simulated text representation generated by the LLM in scenarios where the text modality is missing. $E_{\mathrm{a}}$ refers to the representation obtained from the audio modality and $P$ refers to the input prompt for LLM. The notation ${\rm proj(\cdot)}$ is used to refer to the pre-trained projector network within the SLAM-ASR framework. As shown in the middle of the left side of Fig. \ref{Figure 1.}, ${\rm LLM_g(\cdot)}$ refers to the hidden state derived from the generated results.

\subsubsection{Modality Imagination Autoencoder}
\label{sssec:Modality Imagination Autoencoder}

We use the pre-trained projector and Vicuna to obtain real text representations when text modality is present and generate simulated ones when text modality is missing. However, the gap between real and simulated representations can negatively impact performance if used directly for training. To address this, we use the Residual Autoencoder \cite{Residual_Autoencoder} to construct the missing modality imagination autoencoder, which activates only when the text modality is missing. The structure of MIA is depicted in the bottom left corner of Fig. \ref{Figure 1.}. The definition is as follows: 
\begin{gather}
H_{\mathrm{t}} = f(\bm{W}^{(1)}{\rm cat}(R_{\mathrm{v}}, R_{\mathrm{a}}, \hat{R_{\mathrm{t}}})+\bm{b}^{(1)}) \\
\hat{R_{\mathrm{t}}} = \hat{R_{\mathrm{t}}} + f(\bm{W}^{(2)}H_{\mathrm{t}}+\bm{b}^{(2)})
\end{gather}
$R_{\mathrm{v}}, R_{\mathrm{a}}, \hat{R_{\mathrm{t}}} \in \mathbb{R}^{S\times D}$, where $S$ is the sequence length and $D$ is the unified dimension of modality representations. $H_{\mathrm{t}} \in \mathbb{R}^{S\times D'}$, where $D'$ is the dimension of the intermediate hidden state. We replace the original $\hat{R_{\mathrm{t}}}$ with the reconstructed $\hat{R_{\mathrm{t}}}$. As shown in Fig. \ref{Figure 1.}, since there are two MIA modules, the input is $R_m$ in the MIA-1 module and $R_m^{\mathrm{seq}}$ in the MIA-2 module. 

\subsection{Modality Missing Self-Distillation}
\label{ssec:Modality Missing Self-Distillation}
We design a self-distillation framework to handle cases where all modalities are present and cases where the text modality is missing. The network has two data flows: one with full modalities input (visual, audio, text) and one with only visual and audio modalities input.

During training, we distill text modality representation knowledge from the complete multimodal flow to the scenario where the text modality is missing. To achieve this, we have designed the Modality Knowledge Distillation Loss ($\mathcal{L}_{\mathrm{MKD}}$) and Representation Similarity Loss ($\mathcal{L}_{\mathrm{RS}}$).  To capture the continuous nature of sample orders in the sentiment valence regression task, we introduced the Regression Rank-N Contrast Loss ($\mathcal{L}_{\mathrm{RNC}}$). Additionally, the Mean Squared Error loss is used as the sentiment valence regression task loss ($\mathcal{L}_{\mathrm{Task}}$) for sentiment analysis. To ensure the model acquires these various capabilities, we use the weighted sum of these losses as the final loss function. The overall loss is defined as follows:
\begin{equation}
\mathcal{L}=\mathcal{L}_{\mathrm{Task}}+\alpha \mathcal{L}_{\mathrm{MKD1}} + \beta \mathcal{L}_{\mathrm{MKD2}} + \gamma \mathcal{L}_{\mathrm{RS}} + \delta \mathcal{L}_{\mathrm{RNC}}
\end{equation}
The $\alpha$ to $\delta$ here defines the weights of the various losses, with specific values displayed in our open-sourced code.

\subsubsection{Sentiment Valence Regression Loss}
\label{sssec:Sentiment Valence Regression Task Loss}
This is a regression task for predicting sentiment valence. Thus, the most important loss is the MSE loss between the true and predicted valence, defined as follows: 
\begin{equation}
\mathcal{L}_{\mathrm{Task}} = \frac{1}{n} \sum_{i=1}^{n} (y_{i}-\hat{y_{i}})^2
\end{equation}
Here, $y_{i}$ is the valence label, and $\hat{y_{i}}$ is the predicted valence.

\subsubsection{Modality Knowledge Distillation Loss}
\label{sssec:Modality Knowledge Distillation Loss}
The objective of $\mathcal{L}_{\mathrm{MKD}}$ is to align the simulated text modality representations, produced by the MIA, as similar as possible with the real text modality representations to enhance the model's robustness when text modality is missing. Specifically, we detach the real text modality representation and then compute the RMSE loss between it and the simulated text modality representation from MIA. Consequently, it should be highlighted that the gradient of $\mathcal{L}_{\mathrm{MKD}}$ will only back-propagate through text missing data flow. Considering the presence of two MIA modules, we calculate two MKD losses, which are defined as follows:
\begin{equation}
\mathcal{L}_{\mathrm{MKD}} =  \sqrt{\sum_{i=1}^{N} \frac{(R_{i\mathrm{(detached)}}-\hat{R_{i}})^2}{N}}
\end{equation}
During the process, we need to calculate $\mathcal{L}_{\mathrm{MKD1}}$ and $\mathcal{L}_{\mathrm{MKD2}}$. First, we detach $R_\mathrm{t}$ and use it along with $\hat{R_{\mathrm{t}}}$ in the formula to calculate $\mathcal{L}_{\mathrm{MKD1}}$. Then, we detach $R_{\mathrm{t}}^{\mathrm{seq}}$ and use it along with $\hat{R}_{\mathrm{t}}^{\mathrm{seq}}$ in the formula to calculate $\mathcal{L}_{\mathrm{MKD2}}$.

\subsubsection{Representation Similarity Loss}
\label{sssec:Representation Similarity Loss}
In order to make the representations entering the final regression head as similar as possible when the modality is complete and the text modality is missing, we design a representation similarity loss $\mathcal{L}_{\mathrm{RS}}$.
\begin{equation}
\mathcal{L}_{\mathrm{RS}} =  \sqrt{\sum_{i=1}^{N} \frac{(r_{i}-\hat{r_{i}})^2}{N}}
\end{equation}

\subsubsection{Regression Rank-N Contrast Loss}
\label{sssec:Regression Rank-N Contrast Loss}
For the sentiment valence regression task, the RNC loss has two goals. First, it brings the representations of two data flows of the same data similar. Second, it allows the model to learn and align the representation distance spaces correctly for both the missing text modality and the complete modalities. The RNC loss is defined as follows: 
\begin{gather}
\mathcal{L}_{\mathrm{RNC}}^{i}=\sum _ {j=1,j\neq i}^ {2N} \log \frac {{\rm exp}({\rm sim}(r_ {i},r_ {j})/\tau )}  {\sum _ {r_k \in S_{i,j}} {\rm exp}({\rm sim}(r_ {i},r_ {k})/\tau )}\\
\mathcal{L}_{\mathrm{RNC}}\!=\!\frac {1}{2N}  \sum_{i=1}^ {2N} \frac {-1}{2N-1} \mathcal{L}_{\mathrm{RNC}}^{i} 
\end{gather}
Here, $N$ is the batch size, and $r$ refers to the final representation. The ${\rm sim(\cdot)}$ function computes the L2 distance. $S_{i, j}$ refers to the set of final representations of all other data in the batch whose distance from the valence label of anchor $i$ is greater than or equal to the distance between the labels of $i$ and $j$.

\section{Experiments}
\label{sec:experiments}
\subsection{Dataset and Metric}
\label{ssec:Dataset Analysis}
We test our model on the CMU-MOSEI \cite{CMU-MOSEI} dataset, which includes 22,856 videos: 16,326 for training, 1,871 for
validation, and 4,659 for testing. Each sample is labeled with a sentiment valence from -3 (strongly negative) to +3 (strongly positive). As is common in studies using this dataset, we primarily use Mean Absolute Error (MAE) and prediction Accuracy (ACC) as metrics to evaluate model performance.

\subsection{Implementation Details}
\label{ssec:Implementation details}
In our experiments, we use MANet \cite{MANet} as the visual encoder, WavLM-Large \cite{chen2022wavlm} as the audio encoder, and Vicuna-7B \cite{vicuna2023} as the text encoder. The dimension of the visual representation obtained from the visual encoder is 1024 frame-level features. And we use the output of the 20th hidden layer of WavLM-Large as the audio representation and the output of the last hidden layer to obtain the simulated text representation. When text is present, the sum of the last four hidden layers of Vicuna is used as the text representation. When the text modality is missing, the output of the last hidden layer of the audio encoder is fed into the pre-trained Projector and Vicuna to generate the hidden state, and the penultimate fourth layer of this hidden state is taken as the simulated text representation.

\subsection{Ablation Study}
\label{ssec:Ablation Study}

To evaluate the effectiveness of the missing modality imagination module, the text modality missing self-distillation, and the loss functions, we design a series of ablation experiments to illustrate the role of each component in the architecture. The results are shown in Table \ref{table:1}, where $\mathrm{LLM}_g$ indicates that the LLM-based model generated text representations when the text modality is missing. When ${\rm LLM_g}$ is not used, the model does not use the text modality representations simulated by LLM, but only uses the audio and visual representations for modality imagination and inference.

\begin{table}[h!]

\vspace{-5pt}
\caption{Ablation studies for each module}
\begin{center}
\setlength{\tabcolsep}{1.23mm}
    \begin{tabular}{ccccccccc}
    \hline   
     \multirow{2}*{${\rm LLM_g}$} & \multirow{2}*{MIA} & \multirow{2}*{$\mathcal{L}_{\mathrm{MKD}}$} & \multirow{2}*{$\mathcal{L}_{\mathrm{RS}}$} & \multirow{2}*{$\mathcal{L}_{\mathrm{RNC}}$} & \multicolumn{2}{c}{w/o text} & \multicolumn{2}{c}{w ground truth}\\
     \cmidrule(r){6-7} \cmidrule(r){8-9}
    ~ & ~ & ~ & ~ & ~ & MAE$\downarrow$ & ACC$\uparrow$ & MAE$\downarrow$ & ACC$\uparrow$ \\
    \hline   
     \usym{2717} & \checkmark & \checkmark & \checkmark &\checkmark & 0.584 & 82.47 & 0.522 & 86.57\\
     \checkmark & \usym{2717} & \usym{2717} & \checkmark &\checkmark & 0.572 & 83.81 & 0.521 & 87.45\\
     \checkmark & \usym{2717} & \checkmark & \checkmark &\checkmark & 0.560 & 83.73 & 0.515 & 86.95\\
     \checkmark & \checkmark & \usym{2717} & \checkmark &\checkmark & 0.563 & 83.90 & 0.513 & 87.72\\
     \checkmark & \checkmark & \checkmark & \usym{2717} &\checkmark & 0.552 & 83.84 & 0.510 & 87.53\\
     \checkmark & \checkmark & \checkmark & \checkmark &\usym{2717} & 0.550 & 84.25 & 0.508 & 87.25\\
     \rowcolor{green! 10}\checkmark & \checkmark & \checkmark & \checkmark & \checkmark & \textbf{0.550} & \textbf{84.28} & \textbf{0.506} & 87.64\\
    \hline  
    \end{tabular}
\label{table:1}
\end{center}
\vspace{-5pt}
\end{table}

Using the MAE metric for early stopping, we found that self-distillation significantly improves network performance, with or without the text modality. This demonstrates that our network ensures strong performance even with missing text.

We visualize the similarity matrix between the complete modality representation and the representation with the text modality missing for the same batch of data. The left panel displays the L2 similarity matrix between the complete modality case and the missing text case, arranged by label values. The right panel shows the similarity matrix after removing MIA and distillation losses. Obviously, our model effectively aligns representations under modal integrity and absence, learning the corresponding representation distance based on label distance.

\begin{figure}[t]
  \centering 
  \subfigure[complete]{ 
    \includegraphics[width=1.1in]{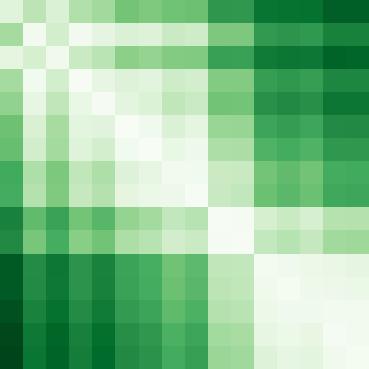}
  } 
  \subfigure[w/o MIA \& $\mathcal{L}_{MKD}$]{ 
    \includegraphics[width=1.1in]{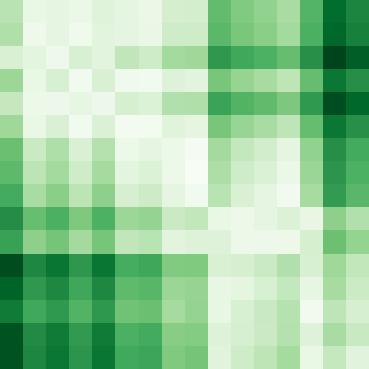}
  } 
  \caption{Feature similarity visualization sorted by labels.} 
\end{figure}




\begin{table}[h!]
\caption{Methods Performance Comparison Table. $\Delta$MAE and $\Delta$ACC represent the absolute difference in model performance between complete modalities and when the text modality is missing. A smaller absolute value indicates better robustness to modality missing.}
\begin{center}
\setlength{\tabcolsep}{1.6mm}
    \begin{tabular}{ccccccc}
    \hline  
     Model & \multicolumn{2}{c}{w text} & \multicolumn{2}{c}{w/o text} & \multicolumn{2}{c}{performance gap}\\
     \cmidrule(r){2-3} \cmidrule(r){4-5} \cmidrule(r){6-7}
     ~ & MAE$\downarrow$ & ACC$\uparrow$ & MAE$\downarrow$ & ACC$\uparrow$ & $\Delta$MAE$\downarrow$ & $\Delta$ACC$\downarrow$\\
     \hline  
     UniMSE \cite{UniMSE} & 0.523 & 87.5 & - & - & - & -\\
     AMB \cite{AMB} & 0.536 & 85.8 & - & - & - & -\\
     ALMT \cite{ALMT} & 0.526 & - & - & - & - & -\\
     MInD \cite{MInD} & 0.529 & - & 0.841 & - & \textcolor{gray}{0.312} & - \\
     GMC \cite{GMC} & - & 80.0 & - & 66.8 & - & \textcolor{gray}{13.2}\\
     GCN \cite{GCN} & - & 86.4 & - & 66.5 & - & \textcolor{gray}{19.9}\\
     MMIN \cite{MMIN} & - & 85.5 & - & 70.1 & - & \textcolor{gray}{15.4} \\
     GMD \cite{GMD} & - & 87.1 & - & 72.6 & - & \textcolor{gray}{14.5}\\

     \rowcolor{green! 10} Ours & \textbf{0.506} & \textbf{87.6} & \textbf{0.550} & \textbf{84.2} & \textbf{0.044} & \textbf{3.4}\\
    \hline 
    \end{tabular}
\label{table:2}
\end{center}
\vspace{-15pt}
\end{table}

\subsection{Overall Comparison}
\label{ssec:Overall Comparison}
We compare our method with other high-performing approaches that have achieved excellent performance or studied modality missing. The approaches listed in Table \ref{table:2} explore robustness against missing modalities using various methods: contrastive learning (GMC), graph neural networks (GCN), modality dependence reduction via gradient-guided decoupling (GMD), missing modality reconstruction (MMIN), and modality information separation (MInD), et al. We referenced the experimental results of these models obtained from articles on the GMD method. The results are shown in Table \ref{table:2}.

First, compared to studies conducted under complete modalities such as UniMSE \cite{UniMSE}, AMB \cite{AMB}, and ALMT \cite{ALMT}, our model achieves significantly better MAE and ACC performance. Second, when the text modality is missing, our model's MAE increases by only 0.044, and the accuracy decreases by just 3.4\%, outperforming other studies on modality-missing robustness. Additionally, the performance decay of our model is an order of magnitude less than that of other models when the text modality is missing, demonstrating its superior design.

\section{Conclusions}
\label{sec:conclusions}
This work addresses the challenge of high cost and heavy reliance on the text modality. Specifically, we designed the
Double-Flow Self-Distillation Framework with Unified Modality Cross-Attention as the main network structure, combined with Modality Imagination Autoencoder to simulate missing text modality. Additionally, we developed a series of loss functions to enhance the model’s performance in multimodal sentiment analysis, ensuring robustness both when all modalities are present and when the text modality is missing.

\section{Acknowledgments}
\label{sec:acknowledgment}

This work was supported by the National Natural Science Foundation of China under Grant 62171427.

\vfill\pagebreak

\bibliographystyle{IEEEtran}
\bibliography{refs}

\end{document}